\pdfoutput=1

\documentclass[11pt]{article}

\usepackage[]{ACL2023}

\usepackage{times}
\usepackage{latexsym}
\usepackage{graphicx}
\usepackage[T1]{fontenc}
\usepackage{subfigure}
\usepackage{tabularx,booktabs}
\usepackage[utf8]{inputenc}
\usepackage{enumitem}
\usepackage{microtype}
\usepackage{amsmath}
\usepackage{inconsolata}
\usepackage{amssymb}
\usepackage{multirow}
\usepackage{algorithm,algorithmic}
\usepackage{pifont}
\newcommand{\xmark}{\ding{55}}%
\newcommand{\cmark}{\ding{51}}%
\newcommand{\ouralg}{TaSL} 

\title{TaSL: Continual Dialog State Tracking via Task Skill
Localization and Consolidation}

\author{Yujie Feng$^1$, Xu Chu$^{2,3,4}$, Yongxin Xu$^{2,3}$, Guangyuan Shi$^1$, \textbf{Bo Liu}$^1$, \textbf{Xiao-Ming Wu}$^1$\thanks{ ~ Corresponding author.} \\
$^1$Department of Computing, The Hong Kong Polytechnic University, Hong Kong S.A.R.\\
$^2$School of Computer Science, Peking University, Beijing, China \\
$^3$Key Laboratory of High Confidence Software Technologies, Ministry of Education, Beijing, China \\
$^4$Center on Frontiers of Computing Studies, Peking University, Beijing, China \\
 yujie.feng@connect.polyu.hk, xiao-ming.wu@polyu.edu.hk 
}

\begin{document}
\maketitle
\begin{abstract}


A practical dialogue system requires the capacity for ongoing skill acquisition and adaptability to new tasks while preserving prior knowledge. However, current methods for Continual Dialogue State Tracking (DST), a crucial function of dialogue systems, struggle with the catastrophic forgetting issue and knowledge transfer between tasks. We present {\ouralg}, a novel framework for task skill localization and consolidation that enables effective knowledge transfer without relying on memory replay. {\ouralg} uses a novel group-wise technique to pinpoint task-specific and task-shared areas. Additionally, a fine-grained skill consolidation strategy protects task-specific knowledge from being forgotten while updating shared knowledge for bi-directional knowledge transfer. As a result, {\ouralg} strikes a balance between preserving previous knowledge and excelling at new tasks. Comprehensive experiments on various backbones highlight the significant performance improvements of {\ouralg} over existing state-of-the-art methods. The source code\footnote{\url{https://github.com/WoodScene/TaSL}} is provided for reproducibility.


\end{abstract}

\section{Introduction}

With the rising popularity of conversational digital assistants, it is imperative for dialogue systems to integrate new services while sustaining proficiency in prior tasks seamlessly. Traditional research, often conducted within specific domains offline, falls short in adaptability to new scenarios~\cite{ni2023recent}. 
Retraining pre-trained language models (PLMs) from scratch is both challenging and resource-intensive~\cite{liu2023good}, highlighting the necessity for efficient continual learning (CL) approaches in dialogue systems~\cite{ke2022continual}. 
Dialogue state tracking (DST), crucial for task-oriented dialogue systems, dynamically updates (domain, slot, value) triplets to capture user intentions precisely. The urgent demand for advancing DST models to accommodate emerging services has catalyzed the development of the Continual DST task~\cite{cho2023continual}.

An effective Continual DST system must address the issue of catastrophic forgetting~\cite{mccloskey1989catastrophic}, where a model's proficiency in old tasks diminishes after learning new ones. It should also promote knowledge transfer (KT)~\cite{ke2021achieving} across domains\footnote{In this work, different \textit{domains} in Continual DST are equivalent to different \textit{tasks} in CL.} to enhance end-task performances. Knowledge transfer includes forward transfer, which improves new task performance using knowledge from previous tasks, and backward transfer, which enhances performance on previous tasks after learning a new relevant task. Striking a balance between retaining previous knowledge and excelling in new tasks is vital for success.


\begin{figure}[t]
  \centering
  \includegraphics[width=1.0\linewidth]{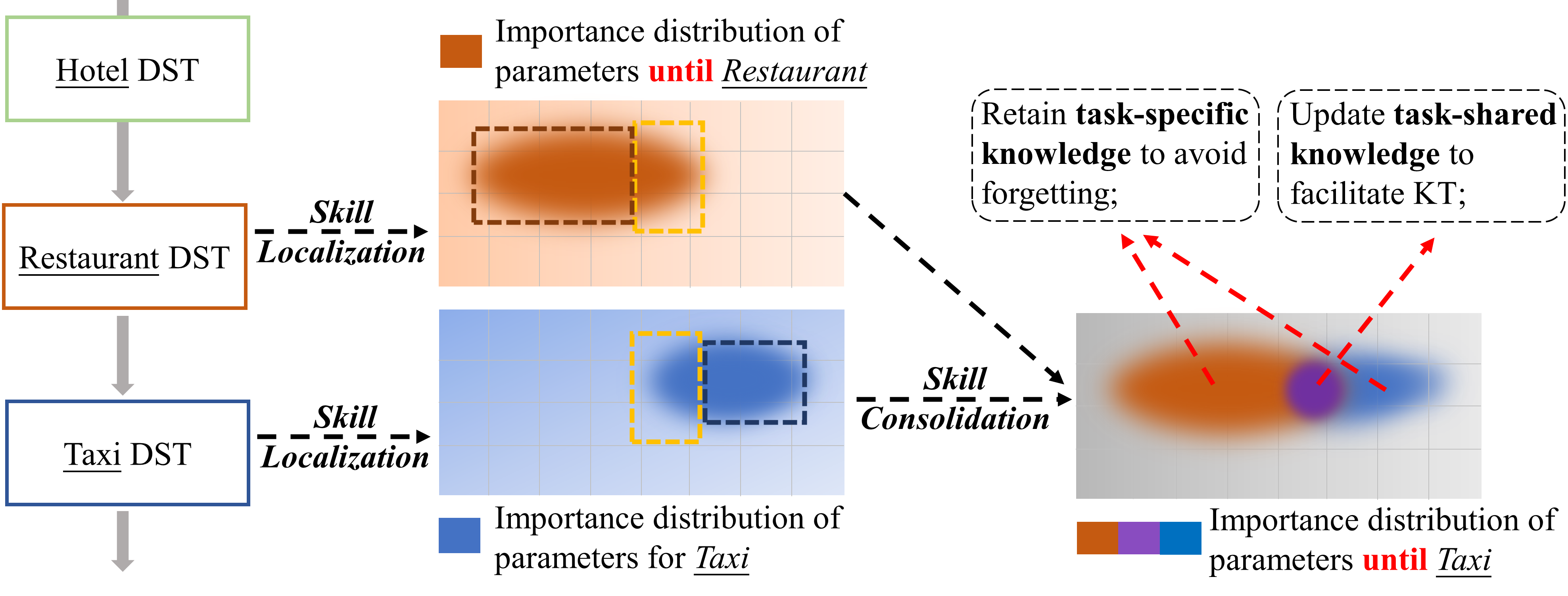}
  \caption{
  Conceptual illustration of \ouralg. By identifying task-relevant areas across both previously accumulated and current tasks, we can consolidate the task-specific and task-shared parameters to facilitate efficient knowledge transfer and mitigate forgetting.
  }
  \label{fig:intro}
\end{figure}

However, current Continual DST methods \cite{madotto2020continual, liu2021domain, cho2023continual, feng2024ros} mainly focus on mitigating forgetting through memory replay or regularization, overlooking the advantages of KT that can be derived from the inherent correlations between different DST domains. 

Task correlation in Continual DST is quite evident. For instance, domains like ``Hotel'' and ``Restaurant'' share semantically similar slots, such as ``area'' and ``bookday'', highlighting the need for models to identify and handle common information types. The similarity in these domain-shared slots is crucial for enabling KT. 
However, learning domain-specific slots like ``food'' for ``Restaurant'' could introduce unique information that disrupts the retention of previously acquired knowledge, leading to catastrophic forgetting.

To address these challenges, we introduce \textbf{Ta}sk \textbf{S}kill \textbf{L}ocalization and Consolidation ({\ouralg}), a framework designed to improve KT between tasks without relying on memory replay. This is achieved by identifying and consolidating the importance distribution of model parameters across tasks. {\ouralg} initially employs a group-wise importance-aware \textit{\textbf{skill localization}} technique that utilizes gradient trajectories to pinpoint tiny regions in the model that store crucial knowledge for the current task. By comparing the importance distribution with those of previous tasks, we can differentiate between task-specific and task-shared parameters, as illustrated in Figure \ref{fig:intro}. 
Our innovative \textit{\textbf{skill consolidation}} phase then categorically integrates weights from previous tasks with the current one, enabling effective KT while minimizing forgetting.

In detail, the importance-aware skill localization method employs a new group-wise metric to compute importance scores, effectively quantifying the significance of each ``\textit{\textbf{skill unit}}''\footnote{The definition of skill units is model-specific. For example, the \textit{Query} matrix within the self-attention layer may be regarded as a skill unit. For detailed definitions, please refer to Appendix \ref{sec:skill_unit}.} for the current task. 
Our approach, focusing on parameter space rather than dataset-driven categorization of domain-shared and domain-specific slots, offers a more robust solution that accurately identifies task-specific and task-shared knowledge, overcoming inaccuracies caused by dataset noise.

Our skill consolidation stage, then based on a fine-grained model averaging strategy, effectively manages different types of knowledge.
We enable forward KT to new tasks by starting with a model initialized with weights from previously fine-tuned tasks, thus using past knowledge to improve learning for new tasks without restrictions. For backward KT, we merge knowledge from both current and past accumulated tasks into localized task-shared skill units, thereby enhancing their capability. To prevent catastrophic forgetting, we consolidate the integrity of skill units containing previous task-specific knowledge, ensuring they remain unaffected by new task learning.
Through extensive experiments on different parameter-level backbones (from 60M to 7B), {\ouralg} exhibits superior performance in mitigating forgetting and showcases remarkable capabilities for KT, significantly outperforming state-of-the-art methods.

Our main contributions include:
\begin{itemize}[leftmargin=*,itemsep=2pt,topsep=0pt,parsep=0pt]
\item 


We propose a novel task skill localization and consolidation
({\ouralg}) \textbf{framework} for CL. 

\item 
We develop new group-wise skill localization and fine-grained skill consolidation \textbf{techniques}. 



\item 

Extensive \textbf{evaluation} on continual DST tasks shows \ouralg{} effectively enables knowledge transfer, resulting in a 3.1\% absolute increase in Avg. JGA and an 8.8\% absolute boost in BWT metrics compared to previous SOTA methods.

\end{itemize}

\section{Related Work}

\subsection{Continual Dialogue State Tracking}
Continual Learning (CL) in task-oriented dialogue systems focuses on perpetually integrating knowledge from data streams for future application.
Three kinds of CL methods have been developed.
Architecture-based methods propose dynamically adding model weights when learning new data \cite{geng2021continual, lu2021getting, yang2023kerprint}.
Replay-based methods store and replay some training samples from previous tasks \cite{hou2019learning, lu2021engage, xu2023seqcare}.
Regularization-based methods employ additional loss functions to solidify new knowledge \cite{li2017learning, xu2023vecocare}.

In the realm of Continual DST, pioneering efforts by \citet{madotto2020continual} and \citet{liu2021domain} have leveraged these CL strategies to set benchmark performance using PLMs. 
The DST-EGQA approach by \citet{cho2023continual} reformulates the DST task to an example-guided question-answering task, aiming to align distribution shifts across different domains to mitigate forgetting. However, these methods overlook DST task correlations that could enhance knowledge transfer.
The recent Continual Prompt Tuning (CPT) method by \citet{zhu2022continual} attempts knowledge transfer via domain-specific soft prompts but depends on inefficient memory replay and extensive retraining.
This dataset-driven approach is inefficient and lacks robustness.

Our {\ouralg} innovates by distinguishing between domain-specific and domain-shared knowledge within the parameter space, then leveraging the skill consolidation process for effective knowledge transfer and forgetting mitigation.

\subsection{Task Skill Localization}
Research indicates that model parameters contribute unevenly to performance \cite{michel2019sixteen}.
\citet{panigrahi2023task} introduced the concept of ``skill localization'' to identify crucial parameters within PLMs, suggesting that fine-tuning critical parameters nearly matches the effect of full fine-tuning. However, their method requires additional time for identifying and retraining key parameters post-fine-tuning, lowering efficiency.

Drawing inspiration from the pruning community, previous studies have used gradient-based metrics to identify important parameters during fine-tuning. Sensitivity-based scoring~\cite{sanh2020movement, zhang2022platon} assesses the impact on training loss and sensitivity smoothing, as applied by \citet{zhang2022adaptive}, eliminates unnecessary parameters for more efficient fine-tuning.
However, these approaches, focusing on individual parameter importance, often lead to element-wise pruning with huge computational and storage burdens~\cite{feng2023towards}. Based on these advances, we introduce a new importance-aware skill localization method, for the first time, that distinguishes between task-specific and shared parameters to mitigate forgetting in CL.




\section{Proposed Method: TaSL}
\paragraph{Problem Formulation}

\begin{figure*}[t]
  \centering
  \includegraphics[width=0.8\linewidth]{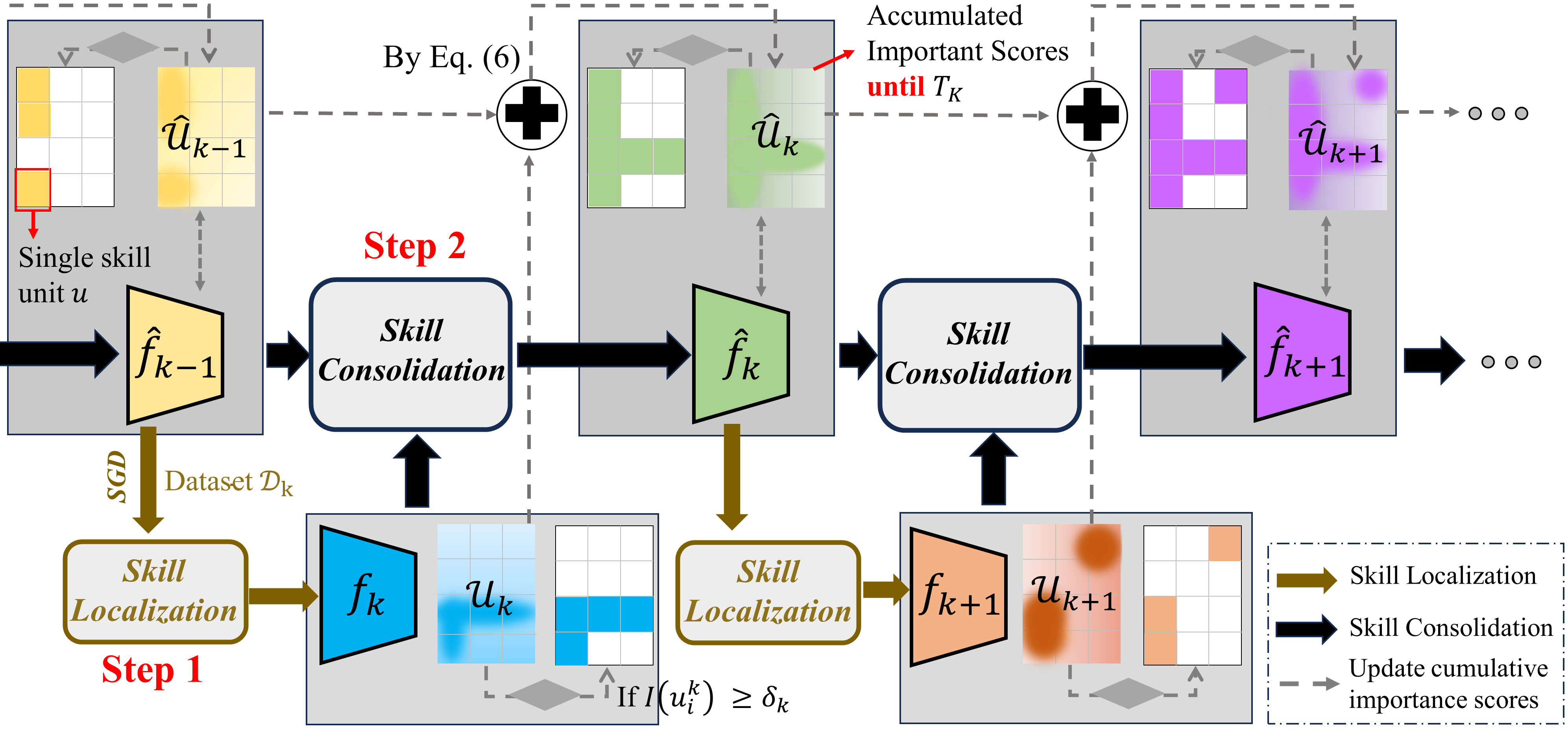}
  \caption{\textbf{Overview of {\ouralg}.}
\textbf{Step 1}: We compute the importance scores of skill units for the current task $\mathcal{T}_{k}$ using our importance-aware skill localization method during fine-tuning.
\textbf{Step 2}: Based on a fine-grained model averaging strategy, the skill consolidation method merges the model $\hat{f}_{k-1}$, which accumulates knowledge of all previous tasks, with the current task's model $f_k$. 
The integration is guided by the importance distributions of skill units across various tasks. We then update the cumulative importance scores for all skill units until task $\mathcal{T}_{k}$ using Eq. (\ref{eq:norm}).
This process is designed to be iteratively repeated with the introduction of each subsequent task.
  }
  \label{fig:method}
\end{figure*}

In continual DST, we aim to train a model $f: \mathcal{X} \times \mathcal{T} \rightarrow \mathcal{Y}$ across a sequence of dialogue domains $\mathcal{T}_1,\ldots,\mathcal{T}_K$.
Each dialogue domain has its dataset $\mathcal{D}_k$ for $\mathcal{T}_k$.
This model predicts the target $y$ based on input $x$ and task $\mathcal{T}_k \in \mathcal{T}$.
The notation $f_k$ refers to the model after training on task $\mathcal{T}_k$, while $\hat{f}_{k}$ denotes the model after averaging for $\hat{f}_{k-1}$ and $f_k$.
Within a given task $\mathcal{T}_k$, a dialogue with $M$ turns of interaction between the system and the user is denoted as $\mathcal{X}_{M} = \left\{\left(A_{1}, U_{1}\right),\left(A_{2}, U_{2}\right) \ldots,\left(A_{M}, U_{M}\right)\right\}$, where $A$ and $U$ represent the system's response and the user's input, respectively.

Each task $\mathcal{T}_k$ is associated with a predefined slot set\footnote{To specify the domain to which a slot belongs, a slot is defined as the concatenation of the specific domain and the slot name, e.g., ``<restaurant-area>''.} $\mathcal{S}=\left\{S_{1}, \ldots, S_{J}\right\}$, where $J$ is the total number of slots.
The objective of DST is to predict the dialogue state $\mathcal{B}_{m}$ based on the dialogue context $\mathcal{X}_{m}$. 
The dialogue state is a collection of (slot, value) pairs, expressed as $\mathcal{B}_{m}=\left\{\left(S_{1}, V_{1}^{m}\right), \ldots,\left(S_{J}, V_{J}^{m}\right)\right\}$, where $V_{J}^{m}$ is the value for slot $S_{J}$ at turn $m$.
DST involves training a model $f: \mathcal{X}_{m} \oplus S_{j} \rightarrow V_{j}^{m}$, with $\oplus$ denotes simple text concatenation.


\paragraph{Overview} {\ouralg} includes two key components: (i) \textit{\textbf{Skill Localization}}, utilizing a new group-wise importance metric to accurately identify the importance distribution of parameters across tasks, and (ii) \textit{\textbf{Skill Consolidation}}, which employs a novel fine-grained model averaging strategy to integrate model weights from both current and past tasks for effective knowledge transfer. Figure \ref{fig:method} provides a comprehensive overview of {\ouralg}, with the following subsections detailing each component.

\subsection{Importance-aware Skill Localization}


To address the substantial computational and storage demands imposed by previous parameter-level importance calculation methods \cite{konishi2023spg}, we propose a new group-wise metric for evaluating the importance of each skill unit $u$:
\begin{equation}
\mathcal{I}(u) = \frac{1}{d_1 \times d_2} \sum\limits_{i=1}^{d_1} \sum\limits_{j=1}^{d_2} s(w_{ij}) \label{eq:3}
\end{equation}
where $w_{ij}$ denotes the trainable parameters, and $d_1 \times d_2$ represents the total parameter count in a skill unit $u$. $\mathcal{I}(u)$ measures the collective importance of all parameters within each skill unit, where higher values signify increased importance.
The function $s(\cdot)$ is a designated importance function for individual parameters, defined as the magnitude of the gradient-weight product:
\begin{equation}
I\left(w_{i j}\right)=\left|w_{i j} \nabla_{w_{i j}} \mathcal{L}\right| \label{eq:1}
\end{equation}

This approximates the loss change when a parameter is zeroed out. If removing a parameter has a significant influence, then the model is sensitive to it, and we should retain it \cite{liang2021super}.

However, sensitivity in Eq. (\ref{eq:1}) may not reliably indicate importance \cite{zhang2022platon}. 
This metric, calculated from a sampled mini-batch, suffers from variability due to stochastic sampling and training dynamics, introducing large uncertainty in estimating sensitivity.
To mitigate this, we apply sensitivity smoothing and uncertainty quantification \cite{zhang2022adaptive}:
\begin{equation}
\begin{split}
\bar{I}^{(t)}\left(w_{i j}\right)  =\alpha_{1} \bar{I}^{(t-1)}\left(w_{i j}\right)+ \left(1-\alpha_{1}\right) I^{(t)}\left(w_{i j}\right) \label{eq:I}
\end{split}
\end{equation}
\begin{equation}
\begin{split}
\bar{U}^{(t)}\left(w_{i j}\right)  =\alpha_{2} \bar{U}^{(t-1)}\left(w_{i j}\right)+ \\ \left(1-\alpha_{2}\right)\left|I^{(t)}\left(w_{i j}\right)-\bar{I}^{(t)}\left(w_{i j}\right)\right| \label{eq:U}
\end{split}
\end{equation}
where $\alpha_{1}$ and $\alpha_{2}$ are smoothing factors, and $t$ is the iteration number.
$\bar{I}^{(t)}$ represents smoothed sensitivity by exponential moving average and $\bar{U}^{(t)}$ is the uncertainty term quantified by the local variation between $I^{(t)}$ and $\bar{I}^{(t)}$.
Importance is then defined by multiplying $\bar{I}^{(t)}$ and $\bar{U}^{(t)}$, providing a more accurate importance assessment for $s(\cdot)$:
\begin{equation}
s^{(t)}\left(w_{i j}\right)=\bar{I}^{(t)}\left(w_{i j}\right) \cdot \bar{U}^{(t)}\left(w_{i j}\right) \label{eq:2}
\end{equation}

\paragraph{Calculating current task importance scores.}
To compute the importance score of each skill unit for task $\mathcal{T}_k$, we employ Eq. (\ref{eq:3}) during fine-tuning. The model $f$ with $n$ skill units is denoted as $\mathcal{U} = \left\{u_1, \ldots, u_n\right\}$, with their importance scores for task $\mathcal{T}_k$ denoted by $\mathcal{I}(\mathcal{U}_k) \in \mathbb{R}^n$.
The detailed computation process is provided in Algorithm~\ref{alg:ipt}.

\paragraph{Computing accumulated importance scores for previous tasks.}
After computing importance scores for each skill unit at the current task $\mathcal{T}_k$, it is essential to compare these with scores from all previously learned tasks to distinguish between task-specific and task-shared parameters. 
To avoid the inefficiency of storing scores for each past task, we aggregate importance scores from all prior tasks into a cumulative score for tasks up to $\mathcal{T}_{k-1}$. This method allows for the iterative refinement of accumulated scores without separately saving past task scores. The skill units with these cumulative scores up to $\mathcal{T}_{k-1}$ are denoted as $\hat{\mathcal{U}}_{k-1}$, calculated using:
\begin{equation}
\begin{split}
\mathcal{I}(\hat{\mathcal{U}}_{k-1}) = \beta \text{Norm} (\mathcal{I}(\hat{\mathcal{U}}_{k-2})) + \\ (1-\beta) \text{Norm} (\mathcal{I}(\mathcal{U}_{k-1})) \label{eq:norm}
\end{split}
\end{equation}
where $\beta \in [0, 1]$, and \text{Norm($\cdot$)} normalizes importance scores to the [0, 1] range, thus resolving discrepancies across models. The initial scores, $\mathcal{I}(\hat{\mathcal{U}}_1)$, are set to be equal to $\mathcal{I}(\mathcal{U}_1)$. Following this, the importance distribution for skill units up to task $\mathcal{T}_{k-1}$ is combined with that of the current task, $\mathcal{T}_{k}$, to facilitate the skill consolidation process.

\subsection{Skill Consolidation}
After skill localization, the subsequent vital phase involves consolidating this knowledge into a unified framework. This process demands a sophisticated model averaging approach considering various factors to optimize task performance.
Traditional coarse-grained model averaging assumes that all model weights are equally important for the training task \cite{kirkpatrick2017overcoming, eddine2023weighted}, which can be written as the following iterative computation format:
\begin{equation}
\hat{f}_k = \lambda \hat{f}_{k-1} + \left(1 - \lambda\right) f_k \label{eq:coarse}
\end{equation}

However, this method may overemphasize weights irrelevant to the current task, contaminating previously acquired task-specific knowledge and leading to forgetting. To counteract this, we introduce a fine-grained averaging strategy focusing on skill units rather than the entire model. Our approach distinguishes between task-shared and task-specific skill units, categorically applying weighted averaging to parameters within each skill unit. 

We initially set importance thresholds $\delta$ using quantiles to select the top 20\% of skill units based on importance scores. A skill unit $u_i^k$ is deemed important (denoted as $(u_i^k)^+$) if its score $\mathcal{I}(u_i^k)$ is above $\delta_k$, and unimportant ($(u_i^k)^-$) otherwise.

\begin{algorithm}[t]
\caption{Importance-aware Skill Localization}\label{alg:ipt}
\begin{algorithmic}[0.9]
\renewcommand{\algorithmicrequire}{\textbf{Input:}}
\renewcommand{\algorithmicensure}{\textbf{Output:}}
\REQUIRE Training dataset $\mathcal{D}_k$ for task $\mathcal{T}_k$; total training iterations $T$; hyperparameters $\alpha_1, \alpha_2$.
\FOR {$t = 1, \ldots, T$}
    \STATE Sample a mini-batch from $\mathcal{D}_k$ and compute the gradient $\nabla \mathcal{L}$;
    \STATE Compute the sensitivity $I\left(w_{i j}\right)$ via Eq. (\ref{eq:1});
    \STATE Update $\bar{I}^{(t)}$ via Eq. (\ref{eq:I}) and $\bar{U}^{(t)}$ via Eq. (\ref{eq:U});
\ENDFOR
\STATE Compute the importance score $\mathcal{I}(u^k_i)$ for each skill unit $u^k_i$ by Eq. (\ref{eq:3}), for $i = 1, \ldots, n$.
\ENSURE  $f_k$ and importance scores $\mathcal{I}(\mathcal{U}_k)$ for $\mathcal{U}_k$.
\end{algorithmic} 
\end{algorithm}

Our fine-grained averaging strategy customizes parameter combination for each skill unit, based on its importance under different tasks, as follows:
\begin{equation}
\hat{u}_i^k =
\begin{cases}
  \gamma \hat{u}_i^{k-1} + (1-\gamma) u_i^k, & \text{if $(\hat{u}_i^{k-1})^+$, $(u_i^k)^+$} \\

  \hat{u}_i^{k-1}, & \text{if $(\hat{u}_i^{k-1})^+$, $(u_i^k)^-$} \\
  
  u_i^k, & \text{if $(\hat{u}_i^{k-1})^-$, $(u_i^k)^+$} \\

  \frac{1}{2}(\hat{u}_i^{k-1} + u_i^k), & \text{if $(\hat{u}_i^{k-1})^-$, $(u_i^k)^-$} \\

\end{cases}
\label{eq:5}
\end{equation}

This strategy performs the element-wise adjustment of parameters within each skill unit based on its relevance to previous and current tasks, using hyperparameter $\gamma$ to control their influences.

In the scenario where a skill unit $u_i$ is significant for both past and present tasks (case 1), we integrate newly acquired knowledge into this task-shared skill unit to enable backward KT. 
If a skill unit $u_i$ is crucial solely for previous tasks (case 2), we maintain the knowledge within this previous task-specific skill unit untouched to prevent the contamination of historical knowledge with task-irrelevant information. 
In contrast, for a skill unit important only to the current task (case 3), since the model $f_k$ is trained on the initialization of $\hat{f}_{k-1}$, the historically learned knowledge is utilized to enhance the performance of the current task, enabling forward KT. Thus, we ensure the integrity of parameters within this current task-specific skill unit, preserving essential knowledge for excelling in the new task. 
We adopt a straightforward averaging for units not pertinent to either task (case 4).

Skill consolidation is performed before starting a new task in CL, utilizing the averaged model for subsequent task initialization. Only the importance scores of $\hat{\mathcal{U}}_{k-1}$ and $\mathcal{U}_k$ are retained for use between tasks, starting with $\hat{\mathcal{U}}_1 = \mathcal{U}_1$ estimated from $f_1$ on $D_1$. 
Detailed implementation of {\ouralg} algorithm can be found in the Appendix (Algorithm~\ref{alg:my_algorithm}).


\section{Experiments and Analysis}\label{sec:exp}
\paragraph{Dataset}
We use the continual learning for DST setup proposed by \citet{zhu2022continual}, which uses 15 single domains from the Schema-Guided Dialog dataset (SGD)~\cite{rastogi2020towards}.
We aggregate our results over the same five domain orders to make the most reliable comparisons with prior works.
Comparing results with the same order is crucial as the results
can have significant variance depending on the chosen domains and their order.
More details about data statistics, task selection, and orderings can be found in the Appendix~\ref{sec:dataset}.

\newcommand{\tabincell}[2]{\begin{tabular}{@{}#1@{}}#2\end{tabular}}
\begin{table*}[t]
\centering
\scalebox{0.9}{
\begin{tabular}{l|c|ccc}
\toprule
Method&   Memory-Free & Avg. JGA&  FWT&  BWT \\
\midrule
\rule{0pt}{4pt} Fine-tuning~\cite{madotto2020continual}  &  \multirow{6}*{\tabincell{c}{\cmark}}   & $44.1_{0.9}$ & $8.3_{1.0}$ & $-36.6_{3.9}$  \\
\rule{0pt}{8pt} EWC~\cite{kirkpatrick2017overcoming}  &   & $47.9_{1.1}$ & $8.4_{0.9}$ & $-38.1_{4.1}$  \\
\rule{0pt}{8pt} AdapterCL~\cite{madotto2020continual}   &    & $49.8_{1.7}$ & - & - \\
\rule{0pt}{8pt} DST-EGQA~\cite{cho2023continual} &    & $55.5_{3.5}$ & $23.6_{2.1}$ & $-19.1_{4.2}$  \\
\rule{0pt}{8pt} RoS~\cite{feng2024ros} &    & $59.0_{3.9}$ & $25.5_{2.0}$ & $-17.9_{3.7}$  \\
\rule{0pt}{8pt} \textbf{{\ouralg} (ours)} &   & $\textbf{62.1}_{2.0}$ & $\textbf{26.6}_{1.5}$ & $\textbf{-9.1}_{2.2}$  \\

\midrule
\rule{0pt}{8pt} Replay~\cite{madotto2020continual}  &  \multirow{4}*{\tabincell{c}{\xmark}}  & $58.6_{3.5}$ & $10.9_{0.5}$ & $-3.2_{2.3}$  \\
\rule{0pt}{8pt} CPT~\cite{zhu2022continual} &    & $61.2_{2.5}$ & $13.7_{0.8}$ & $0.5_{0.4}$ \\
\rule{0pt}{8pt} DST-EGQA~\cite{cho2023continual} &   &  $68.9_{0.3}$ & $22.5_{1.8}$ & $-5.9_{1.9}$  \\
\rule{0pt}{8pt} RoS~\cite{feng2024ros} &   &  $72.1_{0.8}$ & $26.7_{2.0}$ & $-2.6_{1.5}$  \\

\midrule
\rule{0pt}{8pt} CPT Multi-task~\cite{zhu2022continual} & \multirow{3}*{\tabincell{c}{-}}  &  $64.0_{1.9}$ & - & -  \\
\rule{0pt}{8pt} DST-EGQA Multi-task~\cite{cho2023continual} &   & $74.2_{1.8}$ & - & -  \\
\rule{0pt}{8pt} RoS Multi-task~\cite{cho2023continual} &   & $76.3_{0.3}$ & - & -  \\

\bottomrule
\end{tabular}}
\caption{CL results of various methods, all utilizing the same T5-small backbone, on 15 different tasks from the SGD dataset. Means and standard variances are reported across five domain permutations. 
The last two rows provide the multi-tasking results, which serve as an upper bound.
Our memory replay-free {\ouralg} outperforms the previous best method, RoS, by achieving a 3.1\% absolute improvement on avg. JGA and an 8.8\% absolute increase in BWT. Additionally, {\ouralg} exceeds the performance of the majority of memory replay methods and nearly matches the upper bound of the CPT multi-task method.
}
\label{tbl:result}
\end{table*}

\paragraph{Evaluation Protocol}
We evaluate DST performance using the widely adopted Joint Goal Accuracy (JGA) metric \cite{wu2019transferable}, which indicates the percentage
of turns for which all slot values are correctly predicted.
We denote $a_{j,i}$ as the JGA on the test set of task $\mathcal{T}_i$ right after training on task $\mathcal{T}_j$.
The performance of Continual DST is assessed using three metrics from \citet{zhu2022continual}: 
\begin{equation}
    \mathbf{Avg.\ JGA} =\frac{1}{K} \sum_{i=1}^{K} a_{K, i}
\end{equation}
\begin{equation}
    \mathbf{FWT} = \frac{1}{K-1} \sum\limits_{i=2}^{K} a_{i-1, i}
\end{equation}
\begin{equation}
    \mathbf{BWT} = \frac{1}{K-1} \sum\limits_{i=1}^{K-1} a_{K, i}-a_{i, i}
\end{equation}
Avg. JGA represents the average JGA across all tasks after training on the final task $\mathcal{T}_K$.
Forward Transfer (FWT) evaluates a model's generalization ability by measuring the averaged zero-shot performance. 
Backward Transfer (BWT) assesses the impact of learning on subsequent tasks on a previous task.
Negative BWT indicates the model lost some previously acquired knowledge.

\paragraph{Baselines}
We evaluate our method with the following Continual DST baselines:
\textbf{\emph{Fine-tuning:}} Continuously fine-tune the backbone model on new task data.
\textbf{\emph{Replay:}}
Randomly save $\left| \mathcal{M} \right|$ samples from the training set of each previous task $\mathcal{T}_i$ in memory $\mathcal{M}_i$ and jointly train the model on new task data $\mathcal{D}_K$ and memory $\mathcal{M}_{<k}$.
\textbf{\emph{EWC:}}
Maintain a memory but leverage it to compute the Fisher information matrix for regularization \cite{kirkpatrick2017overcoming}.
\textbf{\emph{AdapterCL:}}
Freeze the pre-trained model and independently train a residual Adapter \cite{houlsby2019parameter} for each task \cite{madotto2020continual}.
\textbf{\emph{Continual Prompt Tuning (CPT) \cite{zhu2022continual}:}}
Freeze the backbone model and continually train soft prompts with memory-guided knowledge transfer in both forward and backward directions.
\textbf{\emph{DST-EGQA \cite{cho2023continual}:}}
Reformulate DST as a QA task to mitigate forgetting with retrieval-augmented in-context learning.
\textbf{\emph{RoS \cite{feng2024ros}:}}
Utilize knowledge distillation to enhance the meta-reasoning ability of student models, thereby mitigating forgetting.

\paragraph{Training Details}
We utilize four distinct parameter-level backbones for experiments: T5-small \cite{raffel2020exploring}, T5-base, Flan-T5-large~\cite{chung2022scaling}, and LLaMA-7B \cite{touvron2023llama}. For the T5 series model, we perform full fine-tuning across all parameters. For LLaMA-7B, we adopt the Parameter-Efficient Fine-Tuning technique, specifically Low-Rank Adaptation (LoRA) \cite{hu2021lora}, to expedite the training process.
For {\ouralg}, we set the hyperparameters $\alpha_1$ and $\alpha_2$ in Eq. (\ref{eq:I}) and Eq. (\ref{eq:U}) to 0.85, and set $\beta$ in Eq. (\ref{eq:norm}) and $\gamma$ in Eq. (\ref{eq:5}) to 0.7.
The memory size per task $\left| \mathcal{M} \right|$ is maintained at 50, aligning with previous studies. Detailed training settings are provided in Appendix \ref{sec:skill_unit}.

Following this, we compare {\ouralg} with baselines in Sec. \ref{subsec:exp_main}, and
present a comprehensive ablation study in Sec. \ref{subsec:exp_ablation}.
Subsequently, we delve deeper into the underlying success of our proposed importance-aware skill localization (Sec. \ref{subsec:exp_ipt}) and skill consolidation techniques (Sec. \ref{subsec:exp_avg}), and get some insightful findings from this exploration.

\subsection{Main Results}
\label{subsec:exp_main}
Overall CL results of different methods at the same T5-small backbone are summarized in Table \ref{tbl:result}.

\paragraph{{\ouralg} demonstrates superior CL performance through effective knowledge transfer.}

Unlike vanilla fine-tuning, which suffers from catastrophic forgetting, our approach demonstrates a substantial improvement in Avg. JGA, increasing it from 44.1\% to 62.1\%, and shows marked advancements in both FWT and BWT.

{\ouralg} not only exceeds the CPT method, which relies on memory replay, advancing the Avg. JGA from 61.2\% to 62.1\%, but also establishes a new SOTA across all metrics against the top baseline, DST-EGQA. 
We achieve an impressive increase in Avg. JGA from 59.0\% to 62.1\% (3.1\% absolute improvement) and elevate BWT from -17.9\% to -9.1\%, exceeding it by more than 8\% and displaying robust backward KT capabilities. Additionally, {\ouralg} obtains the highest FWT scores across all baselines under various conditions.

Remarkably, without relying on memory replay, our method nearly matches the performance of DST-EGQA with memory replay, particularly in BWT, with a minimal difference of 3.2\% (-9.1\% vs. -5.9\%). Moreover, our Avg. JGA score closely approaches the upper bound performance set by the CPT multi-task strategy at 64\%, underscoring the effectiveness of our fine-grained model averaging strategy that meticulously accounts for domain-shared and domain-specific parameters.

\begin{figure}[t]
  \centering
  \includegraphics[width=1.0\linewidth]{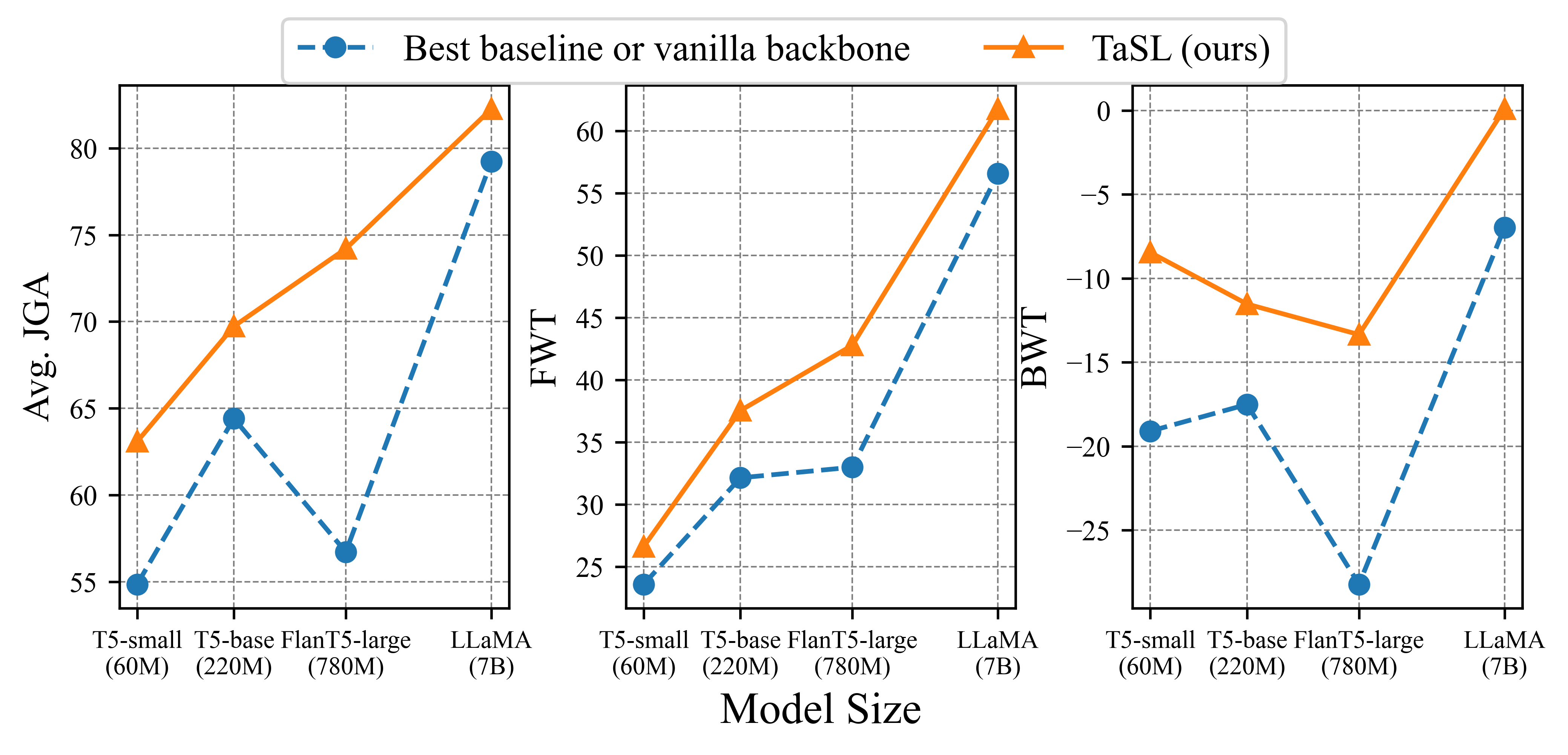}
  \caption{Performance of {\ouralg} w/ different backbones.}
  \label{fig:different_size}
\end{figure}

\paragraph{{\ouralg} consistently demonstrates  superior performance across various backbones.}
To further substantiate the effectiveness of our framework, we conducted experiments using a variety of parameter-level backbones, illustrated in Figure \ref{fig:different_size}, highlighting performance gains with increasing model size. Notably, {\ouralg} achieves a breakthrough by recording a positive BWT score on LLaMA-7B, without employing any memory replay techniques.
Across different backbones, our method consistently outperformed traditional approaches. For instance, in Flan-T5-large, {\ouralg} significantly boosts the Avg. JGA metric from 56\% to 74\%, also achieving the most substantial improvements in both FWT and BWT metrics --- rising from 33\% to 43\% and improving from -28\% to -13\%, respectively.
These results further validate the generality of our proposed framework.

\begin{figure}[t]
  \centering
  \includegraphics[width=0.9\linewidth]{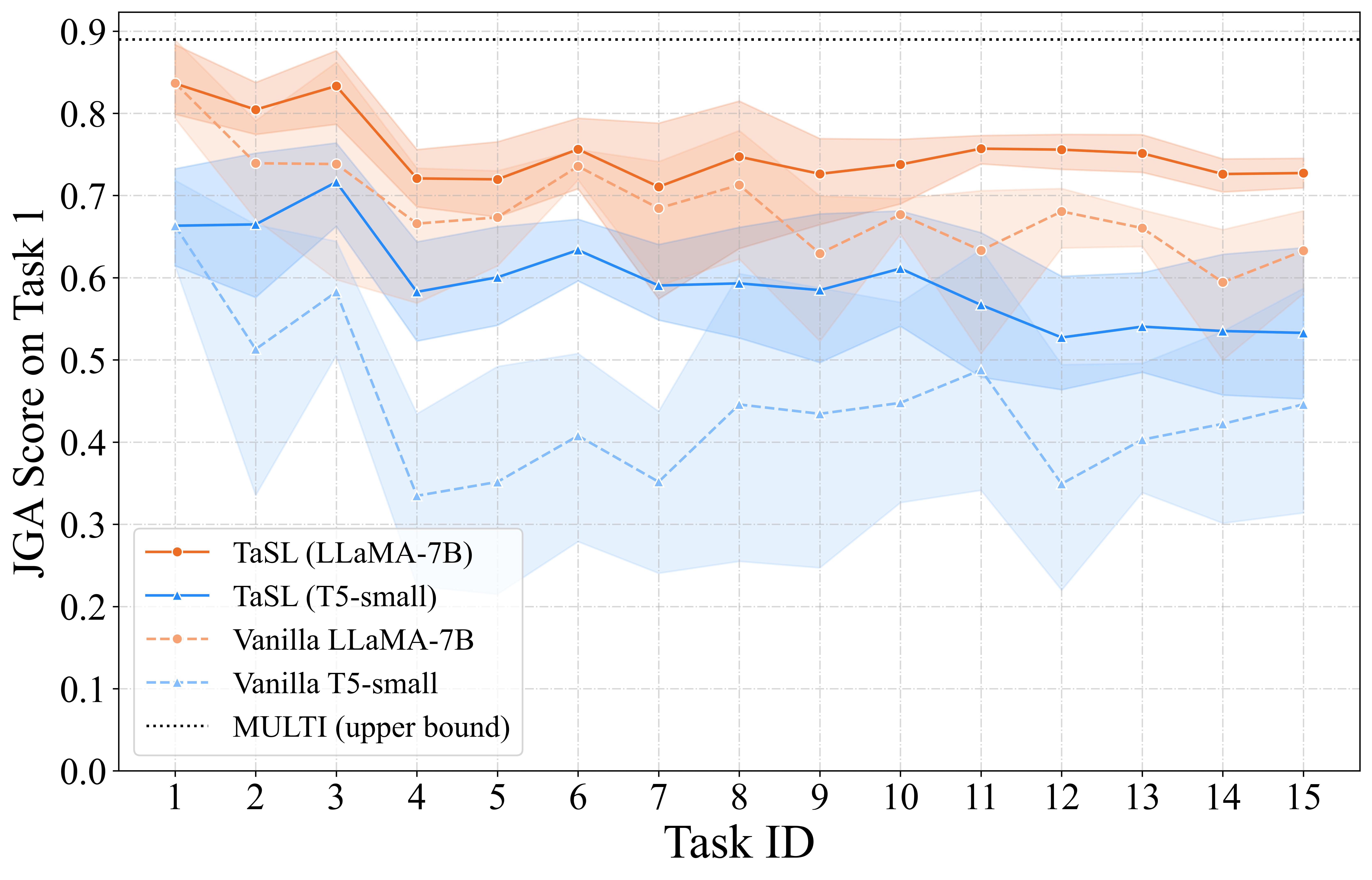}
  \caption{Performance trajectory of Task 1 during the Continual DST learning process.
  }
  \label{fig:forgetting}
\end{figure}

\paragraph{Fine-grained model averaging can effectively mitigate catastrophic forgetting.}
To rigorously evaluate our model's effectiveness in counteracting forgetting, we analyzed its performance on the initial task after training on subsequent tasks. 
Figure \ref{fig:forgetting} illustrates that our method results in a notably slower forgetting rate, manifesting as a nearly 8\% average decrease in performance after training on the last task. 
This contrasts sharply with vanilla backbones, which display a substantial performance reduction of 20\% on average, thereby underscoring our method's superior capacity to mitigate forgetting. Moreover, an intriguing observation is the enhancement in performance on task 1 after training on task 3, highlighting our model's effective backward KT ability.

\subsection{Ablation Study}
\label{subsec:exp_ablation}


This section we assess the effects of importance-aware skill localization and fine-grained skill consolidation, with the results discussed below. For hyperparameter sensitivity, see Appendix \ref{sec:hyper}.

\paragraph{Various importance scoring methods for skill localization.}
Our method calculates importance scores by Eq. (\ref{eq:3}). As shown in Table \ref{tbl:ablation_ipt}, we explore alternative importance scoring approaches: (i) modifying $s(\cdot)$ in Eq. (\ref{eq:3}) only to include sensitivity, as in Eq. (\ref{eq:1}); and (ii) using absolute gradients, $\left|\nabla_{w_{ij}} \mathcal{L}\right|$, for importance assessment~\cite{michel2019sixteen}.
The results indicate that using moving averages for importance scoring outperforms the alternatives, with the other two variants leading to a maximum performance decrease by 3.26\%, 2.24\%, and 4.11\% across the three metrics. This highlights the value of accurate skill localization in improving model performance.

\begin{table}
\centering
\scalebox{0.9}{
\begin{tabular}{lccc}
\toprule
Method & Avg. JGA & FWT & BWT\\
\midrule
\rule{0pt}{6pt} vanilla T5-small &44.10 & 8.32 & -36.63 \\
\midrule
\rule{0pt}{8pt} $s\left( \cdot \right) = I\left( \cdot \right)$ &  60.48 & 24.39  & -10.81 \\
\rule{0pt}{8pt} $s\left( \cdot \right) = \left|\nabla_{w_{i j}} \mathcal{L}\right|$ & 58.82  &  24.80 & -13.22 \\
\rule{0pt}{8pt} {\ouralg} (ours) & \textbf{62.08}  & \textbf{26.63}  & \textbf{-9.11} \\

\bottomrule
\end{tabular}}
\caption{Ablation study. Evaluating the impact of importance scoring variations on skill localization.
}
\label{tbl:ablation_ipt}
\end{table}


\paragraph{Fine-grained vs. coarse-grained model averaging for skill consolidation.}
We compared our fine-grained averaging strategy against two coarse-grained strategies: (i) Weight-Ensemble, which averages weights uniformly as per Eq. (\ref{eq:coarse}), and (ii) Exponential Moving Average (EMA) \cite{szegedy2016rethinking}, applying a running average of parameters at each fine-tuning iteration. Results are detailed in Table \ref{tbl:ablation_avg}.

\begin{table}[t]
\centering
\scalebox{0.8}{
\begin{tabular}{lccc}
\toprule
Method & Avg. JGA & FWT & BWT\\
\midrule
\rule{0pt}{6pt} vanilla T5-small &44.10 & 8.32& -36.63 \\
\midrule
\rule{0pt}{8pt} Weight-Ens.& 53.23  & 21.73  & -18.28 \\
\rule{0pt}{8pt} EMA & 52.56  & 22.27  &  -16.80\\
\rule{0pt}{8pt} Fine-grained (ours) & \textbf{62.08}  & \textbf{26.63}  & \textbf{-9.11} \\
\bottomrule
\end{tabular}}
\caption{Ablation study. Comparing coarse- and fine-grained model averaging methods on skill consolidation.}
\label{tbl:ablation_avg}
\end{table}

Weight-Ensemble significantly improves upon the vanilla model, highlighting coarse-grained averaging's benefits for Continual DST. EMA generally surpasses Weight-Ensemble but falls short of our fine-grained approach due to its overuse of averaging, with frequent parameter adjustments within the same task possibly resulting in less optimal outcomes. Our method solely averages weights after each task, enhancing computational efficiency. 

\subsection{Visualization of Skill Units}
\label{subsec:exp_ipt}

\begin{figure}[t]
  \centering
  \includegraphics[width=1\linewidth]{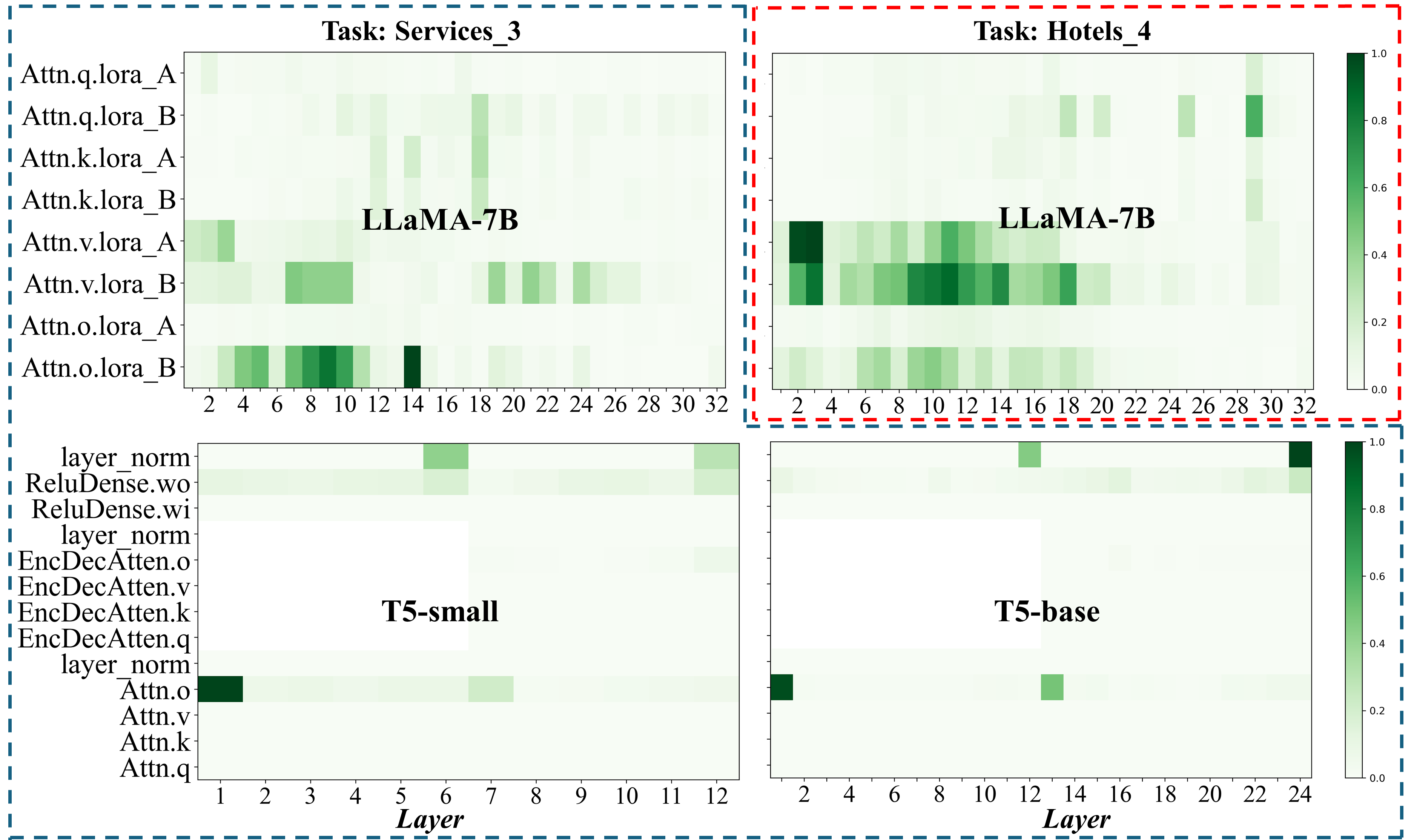}
  \caption{Visualization of importance scores for skill units across different backbone models and tasks.
  }
  \label{fig:heatmap}
\end{figure}

We visualized the distribution of importance scores for the skill units across tasks and models, as shown in Figure \ref{fig:heatmap}, leading to several critical insights:

\begin{itemize}[leftmargin=*,itemsep=2pt,topsep=0pt,parsep=0pt]
\item 

There is a noticeable variation in the importance of skill units for the same task, with important skill units in LLaMA-7B making up about 20\% of all trainable LoRA parameters.

\item 

The distribution of important skill units is task-dependent, indicating both task-shared and specific parameters, confirming {\ouralg}'s validity.

\item 

Lower layers, nearer to the input, are more crucial for the DST task compared to upper layers.

\item 

Within each layer, the importance of the attention layer, especially the V (value) and O (output) matrices, consistently exceeds that of the Q (query), K (key) matrices, and the MLP layer.

\end{itemize}

\subsection{Improved Balance in Knowledge Transfer}
\label{subsec:exp_avg}


This section evaluates the effectiveness of our fine-grained model averaging method in achieving the optimal balance between preserving previous knowledge and excelling at new tasks in CL, comparing it to coarse-grained approaches.

Table \ref{tbl:balance} shows that for a sequence of two tasks, vanilla fine-tuning on LLaMA-7B results in a notable decline in historical Task 1's performance (from 92.3\% to 73.1\%), indicating notable forgetting. Coarse-grained averaging mitigates this to an extent, reducing the decline to 82.3\% but impacting new task performance to 71.3\%. Our method effectively lessens forgetting (improving to 86.9\%) while also maintaining better performance on Task 2, with less than a 3\% reduction.

As tasks increase to three, our method more effectively compensates for losses on new tasks with gains on historical tasks. Vanilla fine-tuning on T5-small results in a combined 55.6\% drop in Tasks 1 and 2, while our approach only shows a 16.6\% decrease and the loss on task 3 is less than 1\% due to {\ouralg}'s effective KT ability.


\begin{table}[t]
\centering
\scalebox{0.8}{
\begin{tabular}{clccc}
\toprule
Seq. length & Method & Task1 &  Task2  &Task3  \\
\midrule

\rule{0pt}{4pt}\multirow{2}*{\tabincell{c}{\\ 2\\(LLaMA-7B)}}   & Upper bound&92.3 &86.1 & -    \\

\rule{0pt}{8pt}  & Fine-tuning &73.1 &86.1 & -    \\
\rule{0pt}{8pt}  & Coarse-grained  & 82.3 & 71.3 & -    \\
\rule{0pt}{8pt}  & {\ouralg} (ours)  &\textbf{86.9}  &\textbf{83.3} &-   \\
\midrule
\rule{0pt}{8pt}\multirow{2}*{\tabincell{c}{\\ 3\\(T5-small)}} & Upper bound & 89.2 & 81.5 &  64.4\\

\rule{0pt}{8pt}  & Fine-tuning  & 53.1 &62.0 &64.4 \\
\rule{0pt}{8pt}  & Coarse-grained  &  58.5 &  64.6 & 43.7    \\
\rule{0pt}{8pt}  & {\ouralg} (ours)   &\textbf{80.0}  &\textbf{74.1} & \textbf{63.8}   \\
\bottomrule
\end{tabular}}
\caption{Analysis of knowledge balancing across old and new tasks. All results are reported in JGA(\%).}
\label{tbl:balance}
\end{table}

\section{Conclusion}

In this paper, we introduce a novel {\ouralg} method to enhance Continual DST performance by facilitating effective knowledge transfer across tasks. 
Our approach leverages an innovative importance-aware skill localization technique and a skill consolidation strategy to differentiate between domain-specific and domain-shared parameters, mitigating forgetting. 
Comprehensive experiments showcase our method's exceptional ability to balance preserving past knowledge and excelling in new tasks.

\section*{Limitations}

{\ouralg} excels at precisely distinguishing between task-specific and shared parameters through importance-aware skill localization. However, the current importance scoring criteria, relying on first-order gradients, may lack precision. The Hessian matrix often captures the actual importance, but computing these second-order gradients incurs significant computational costs. Therefore, future improvements should focus on developing more accurate and efficient skill localization methods.

In addition, in skill consolidation, the challenge lies in better integrating model parameters. Under our fine-grained model averaging strategy (Eq. (\ref{eq:5})), selecting different weighted combinations could impact the overall performance. Although we investigated the model's sensitivity to various hyperparameter settings (Appendix \ref{sec:hyper}), with results showing stable and consistently strong performance across different combinations, there is still potential for further improvement. For instance, developing adaptive methods to select optimal weights or devising more efficient model averaging strategies could further enhance model performance.

\section*{Acknowledgments}
We thank the anonymous reviewers for their valuable feedback. This research was partially supported by the grant of HK ITF ITS/359/21FP.

\bibliography{acl2023}
\bibliographystyle{acl_natbib}

\appendix

\section{Dataset Statistics}
\label{sec:dataset}
Here, we offer a detailed description of the dataset used in Continual DST. Table \ref{tab:dataset_count} displays the number of slots for each of the 15 services used in our experiments and the count of samples in the training, validation, and test sets. Table \ref{tab:dataset_order} illustrates the training sequence for these 15 tasks in the context of continual learning.

\begin{table*}[t]
\centering
\scalebox{0.9}{
\begin{tabular}{ccrrrrrrrrr}
\toprule
Task ID &
  Service &
  \multicolumn{1}{c}{\# Slots} &
  \multicolumn{3}{c}{\# Dialogs} &
  \multicolumn{3}{c}{\# Samples} &
  \multicolumn{2}{c}{Avg. tokens} \\ \midrule
\textit{} &
  \textit{} &
  \multicolumn{1}{c|}{\textit{}} &
  \multicolumn{1}{c}{\textit{Train}} &
  \multicolumn{1}{c}{\textit{Dev}} &
  \multicolumn{1}{c|}{\textit{Test}} &
  \multicolumn{1}{c}{\textit{Train}} &
  \multicolumn{1}{c}{\textit{Dev}} &
  \multicolumn{1}{c|}{\textit{Test}} &
  \multicolumn{1}{c}{\textit{Context}} &
  \multicolumn{1}{c}{\textit{Query}} \\ \midrule
30 & services\_4    & \multicolumn{1}{c|}{5}  & 86  & 13 & \multicolumn{1}{r|}{25}  & 680  & 97  & \multicolumn{1}{r|}{208}  & 154 & 49 \\
31 & flights\_1     & \multicolumn{1}{c|}{10} & 560 & 80 & \multicolumn{1}{r|}{160} & 4680 & 667 & \multicolumn{1}{r|}{1379} & 168 & 10 \\
32 & services\_3    & \multicolumn{1}{c|}{5}  & 131 & 19 & \multicolumn{1}{r|}{38}  & 959  & 143 & \multicolumn{1}{r|}{290}  & 143 & 54 \\
33 & flights\_3     & \multicolumn{1}{c|}{8}  & 65  & 10 & \multicolumn{1}{r|}{19}  & 420  & 75  & \multicolumn{1}{r|}{116}  & 133 & 79 \\
34 & trains\_1      & \multicolumn{1}{c|}{7}  & 58  & 9  & \multicolumn{1}{r|}{17}  & 415  & 67  & \multicolumn{1}{r|}{117}  & 131 & 76 \\
35 & homes\_2       & \multicolumn{1}{c|}{8}  & 62  & 9  & \multicolumn{1}{r|}{18}  & 424  & 56  & \multicolumn{1}{r|}{139}  & 140 & 89 \\
36 & rentalcars\_2  & \multicolumn{1}{c|}{6}  & 77  & 11 & \multicolumn{1}{r|}{23}  & 631  & 91  & \multicolumn{1}{r|}{185}  & 157 & 61 \\
37 & restaurants\_1 & \multicolumn{1}{c|}{9}  & 256 & 37 & \multicolumn{1}{r|}{74}  & 2098 & 297 & \multicolumn{1}{r|}{581}  & 153 & 10 \\
38 & music\_1       & \multicolumn{1}{c|}{6}  & 68  & 10 & \multicolumn{1}{r|}{20}  & 468  & 73  & \multicolumn{1}{r|}{142}  & 118 & 61 \\
39 & hotels\_4      & \multicolumn{1}{c|}{7}  & 80  & 12 & \multicolumn{1}{r|}{23}  & 559  & 99  & \multicolumn{1}{r|}{141}  & 134 & 72 \\
40 & media\_2       & \multicolumn{1}{c|}{5}  & 32  & 4  & \multicolumn{1}{r|}{10}  & 215  & 29  & \multicolumn{1}{r|}{71}   & 112 & 59 \\
41 & hotels\_3      & \multicolumn{1}{c|}{6}  & 90  & 13 & \multicolumn{1}{r|}{26}  & 737  & 100 & \multicolumn{1}{r|}{193}  & 157 & 64 \\
42 & rentalcars\_3  & \multicolumn{1}{c|}{7}  & 44  & 7  & \multicolumn{1}{r|}{13}  & 332  & 55  & \multicolumn{1}{r|}{99}   & 148 & 72 \\
43 & hotels\_1      & \multicolumn{1}{c|}{7}  & 99  & 14 & \multicolumn{1}{r|}{29}  & 868  & 105 & \multicolumn{1}{r|}{250}  & 161 & 71 \\
44 & homes\_1       & \multicolumn{1}{c|}{7}  & 244 & 35 & \multicolumn{1}{r|}{70}  & 1829 & 282 & \multicolumn{1}{r|}{540}  & 159 & 81  \\ 
\bottomrule
\end{tabular}}%
\caption{Statistics of the 15 services we used in experiments. 
}
\label{tab:dataset_count}
\end{table*}

\begin{table*}[t]
\centering
\begin{tabular}{c|ccccccccccccccc}
\toprule
Task order & \multicolumn{15}{c}{Tasks' IDs in order} \\
\midrule
Order1 & 30 & 31 & 32 & 33 & 34 & 35 & 36 & 37 & 38 & 39 & 40 & 41 & 42 & 43 & 44 \\
Order2 & 39 & 33 & 36 & 42 & 40 & 37 & 38 & 34 & 32 & 35 & 41 & 31 & 30 & 44 & 43 \\
Order3 & 30 & 41 & 38 & 31 & 43 & 39 & 40 & 33 & 34 & 44 & 37 & 36 & 32 & 35 & 42 \\
Order4 & 43 & 40 & 44 & 38 & 30 & 37 & 31 & 39 & 32 & 35 & 41 & 34 & 33 & 36 & 42 \\
Order5 & 30 & 33 & 44 & 31 & 38 & 32 & 42 & 40 & 37 & 43 & 36 & 39 & 41 & 35 & 34 \\ \bottomrule
\end{tabular}%
\caption{Five task orders of all our 15 tasks experiments.
}
\label{tab:dataset_order}
\end{table*}

\section{Implementation Details}
\label{sec:skill_unit}
\paragraph{Definition of Skill Units}
In our Importance-aware Skill Localization technique, to circumvent the high computational resource demands of parameter-level localization, we introduced a novel group-wise metric, redefining the skill unit $u$ as the basic element for computing importance scores. However, the division of skill units varies across different backbone models due to variations in parameter quantities and architectural designs (e.g., decoder-only and encoder-decoder architectures). The specific distinctions are as follows:
\begin{itemize}[leftmargin=*,itemsep=2pt,topsep=0pt,parsep=0pt]
\item \textbf{Encoder-decoder architecture backbones.}
For these backbones~\cite{feng2021completing}, such as T5-small, T5-base, and T5-large, we perform full-parameter fine-tuning during training. For organizational simplicity, we divided the models based on the different functionalities within the transformer blocks, as depicted in Table \ref{tbl:en-de}. For instance, in T5-small, with both encoder and decoder comprising 6 transformer blocks each, the total comes to 131 skill units for T5-small, 257 skill units for T5-base, and 558 for T5-large.

\begin{table}[h]
\centering
\scalebox{0.75}{
\begin{tabular}{ll}
\toprule
Block Type &   Skill Unit Name\\
\midrule
\rule{0pt}{4pt}\multirow{8}*{\tabincell{c}{\\ Encoder}}  & SelfAttention.q.weight \\
\rule{0pt}{8pt}  & SelfAttention.k.weight \\
\rule{0pt}{8pt}  & SelfAttention.v.weight \\
\rule{0pt}{8pt}  & SelfAttention.0.weight \\
\rule{0pt}{8pt}  & layer.0.layer\_norm.weight \\
\rule{0pt}{8pt}  & DenseReluDense.wi.weight \\
\rule{0pt}{8pt}  & DenseReluDense.wo.weight \\
\rule{0pt}{8pt}  & layer.1.layer\_norm.weight \\
\midrule
\rule{0pt}{8pt}\multirow{11}*{\tabincell{c}{\\ Decoder}}& SelfAttention.q.weight  \\
\rule{0pt}{8pt}  & SelfAttention.k.weight \\
\rule{0pt}{8pt}  & SelfAttention.v.weight \\
\rule{0pt}{8pt}  & SelfAttention.0.weight \\
\rule{0pt}{8pt}  & SelfAttention.relative\_attention\_bias.weight \\
\rule{0pt}{8pt}  & layer.0.layer\_norm.weight \\
\rule{0pt}{8pt}  & layer.1.EncDecAttention.q.weight \\
\rule{0pt}{8pt}  & layer.1.EncDecAttention.k.weight \\
\rule{0pt}{8pt}  & layer.1.EncDecAttention.v.weight \\
\rule{0pt}{8pt}  & layer.1.EncDecAttention.o.weight \\
\rule{0pt}{8pt}  & 1.layer\_norm.weight \\
\bottomrule
\end{tabular}}
\caption{Definition of skill units for encoder-decoder architecture backbones at each transformer block.}
\label{tbl:en-de}
\end{table}

\item \textbf{Decoder-only architecture backbone.}
The LLaMA-7B model we utilized falls into this category~\cite{feng2023towards2}. Leveraging Parameter-efficient Fine-tuning Techniques (PEFT) to expedite training, we treat the matrices A and B in LoRA as individual basic skill units. And we add LoRA adapters to attention layers in LLaMA. Each layer, as detailed in Table \ref{tbl:de-on}, comprises 8 distinct skill units. Given that LLaMA-7B consists of 32 layers, it is thereby segmented into 256 skill units in total.

\begin{table}[h]
\centering
\scalebox{0.85}{
\begin{tabular}{ll}
\toprule
Block Type &   Skill Unit Name\\
\midrule
\rule{0pt}{4pt}\multirow{8}*{\tabincell{c}{\\ Decoder}}  & self\_attn.q\_proj.lora\_A.default.weight \\
\rule{0pt}{8pt}  & self\_attn.q\_proj.lora\_B.default.weight \\
\rule{0pt}{8pt}  & self\_attn.k\_proj.lora\_A.default.weight \\
\rule{0pt}{8pt}  & self\_attn.k\_proj.lora\_B.default.weight \\
\rule{0pt}{8pt}  & self\_attn.v\_proj.lora\_A.default.weight \\
\rule{0pt}{8pt}  & self\_attn.v\_proj.lora\_B.default.weight \\
\rule{0pt}{8pt}  & self\_attn.o\_proj.lora\_A.default.weight \\
\rule{0pt}{8pt}  & self\_attn.o\_proj.lora\_B.default.weight \\
\bottomrule
\end{tabular}}
\caption{Definition of skill units for decoder-only architecture backbones at each transformer block.}
\label{tbl:de-on}
\end{table}

\end{itemize}

\paragraph{Model training details}
For different backbones, we utilized the following hyperparameters:

\begin{itemize}[leftmargin=*,itemsep=2pt,topsep=0pt,parsep=0pt]
\item \textbf{T5-small (60M)}, \textbf{T5-base (220M)}, and \textbf{FLAN-T5-lare (780M)}: Training was conducted with a learning rate of 3e-4, batch size of 8, maximum input length of 512, maximum target length of 128, and 5 epochs.
\item \textbf{LLaMA (7B)}: Utilizing LORA for efficiency, with a learning rate of 3e-4, batch size of 128, a cutoff length of 512, and 5 epochs. Lora settings were r = 8, alpha = 16, dropout = 0.05, targeting modules [[q\_proj,k\_proj,v\_proj,o\_proj]]. For testing, settings included temperature = 0.02, top\_p = 0, top\_k = 1, num\_beams = 1, max new tokens = 128.
\end{itemize}

Experiments are carried out using 2 NVIDIA A100 with 80GB memory.  Results are averaged across five different task orders and include the standard error in the tables and plots provided~\cite{feng2022spatial}.

\section{Additional Results}
To further validate TaSL's effectiveness in more complex continual learning scenarios, we have conducted additional experiments to verify its performance on transitioning between different datasets, specifically from SGD to MultiWoz. This involved including another 5 distinct domains from the MultiWoz 2.1 dataset, in addition to the 15 domains in SGD, resulting in a total of 20 domains (i.e., tasks). Table \ref{tbl:diff-dataset} presents the performance of various methods, utilizing T5-small as the backbone.

\begin{table}[h]
\centering
\scalebox{0.75}{
\begin{tabular}{lcccc}
\toprule
Method &   Memory-Free	 & avg. JGA   & FWT & BWT\\
\midrule

\rule{0pt}{4pt}Fine-tuning  &  \multirow{3}*{\tabincell{c}{\cmark}}	  &	20.1&  6.6  & -53.1\\
\rule{0pt}{8pt}DST-EGQA  & 	  &	40.5&  18.4  & -37.1\\
\rule{0pt}{8pt}TaSL (ours)  & 	  &	\textbf{49.9}&  \textbf{22.0}  & \textbf{-23.8}\\
\midrule
\rule{0pt}{8pt}Replay  & \multirow{2}*{\tabincell{c}{\xmark}} 	  &	47.2&  7.3  & -16.0\\
\rule{0pt}{8pt}DST-EGQA  & 	  &	51.2& 18.5  & -21.9\\

\bottomrule
\end{tabular}}
\caption{Cross-dataset performance of TaSL.}
\label{tbl:diff-dataset}
\end{table}

The findings align with the observations from Table \ref{tbl:result}, although there is a noticeable decrease in the efficacy of all evaluated methods, due to the significant discrepancies in data distribution across the datasets examined. As a memory-free method, our TaSL still significantly outperforms the strongest baseline (i.e., the memory-free version of DST-EGQA) and even surpasses some memory-based methods like “Replay”. These findings demonstrate TaSL's robustness and effectiveness, showcasing its capability to handle complex continual learning scenarios.

\section{Sensitivity Analysis for Hyperparameters}
\label{sec:hyper}
The proposed framework incorporates three key hyperparameters, including the $\alpha$ for computing importance scores in Equations (\ref{eq:I}) and (\ref{eq:U}), the $\beta$ for calculating cumulative importance scores in Equation (\ref{eq:norm}), and the $\gamma$ for performing weighted averaging within skill units as outlined in Equation (\ref{eq:5}). Our analysis aims to assess the impact of varying these hyperparameters on our method's performance, testing on the T5-small backbone model.

As evidenced in Table \ref{tbl:hp_alpha}, we determine that the optimal setting for $\alpha$ is 0.55. An $\alpha$ value too low results in a performance decline, indicating that the calculated importance scores are not sufficiently accurate. Furthermore, as depicted in the results of Tables \ref{tbl:hp_beta} and \ref{tbl:hp_gamma}, we also find that $\beta$ and $\gamma$ values within a normal range do not significantly affect performance. However, excessively high or low values for $\beta$ and $\gamma$ may skew the model towards favoring either past or current task knowledge, thereby disrupting the desired balance. Nonetheless, the model's performance remains relatively stable across most conditions, indicating a low sensitivity to hyperparameter variations.

\begin{table}[t]
\centering
\scalebox{1}{
\begin{tabular}{lllr}
\toprule
$\alpha_1$,$\alpha_2$ &   avg. JGA	& FWT   & BWT \\
\midrule
\rule{0pt}{4pt} \textit{fine-tuning}  & 41.6 & 9.6 & -36.7 \\
\midrule
\rule{0pt}{8pt} 0.15  & 61.8  &  29.7 & -10.7\\
\rule{0pt}{8pt} 0.35 &  61.2 & 30.1	   &-12.3\\
\rule{0pt}{8pt} 0.55 &	62.8 & 28.6	   & -9.5\\
\rule{0pt}{8pt} 0.85 &	60.7 & 28.9	   &-10.3\\
\rule{0pt}{8pt} 0.95 &	61.7 & 30.0	   &-10.6\\

\bottomrule
\end{tabular}}
\caption{Performance comparisons of {\ouralg} (using T5-small as
the backbone) equipped with different $\alpha$ at task order 1.}
\label{tbl:hp_alpha}
\end{table}

\begin{table}[t]
\centering
\scalebox{1}{
\begin{tabular}{lllr}
\toprule
$\beta$ &   avg. JGA	& FWT   & BWT \\
\midrule
\rule{0pt}{4pt} \textit{fine-tuning}  & 41.6 & 9.6 & -36.7 \\
\midrule
\rule{0pt}{8pt} 0.1  & 61.8  &  29.6 &-11.7\\
\rule{0pt}{8pt} 0.3 &	61.5& 28.4	   &-11.4\\
\rule{0pt}{8pt} 0.5 &	62.3& 	 29.5  &-10.2\\
\rule{0pt}{8pt} 0.7 &	60.7& 	 28.9  &-10.3\\
\rule{0pt}{8pt} 0.9 &	58.2& 	 30.2  &-13.0\\
\bottomrule
\end{tabular}}
\caption{Performance comparisons of {\ouralg} (using T5-small as
the backbone) equipped with different $\beta$ at task order 1.}
\label{tbl:hp_beta}
\end{table}

\begin{table}[t]
\centering
\scalebox{1}{
\begin{tabular}{lllr}
\toprule
$\gamma$ &   avg. JGA	& FWT   & BWT \\
\midrule
\rule{0pt}{4pt} \textit{fine-tuning}  & 41.6 & 9.6 & -36.7 \\
\midrule
\rule{0pt}{8pt} 0.1  & 60.1 & 28.2 &-12.1\\
\rule{0pt}{8pt} 0.3 &	61.7& 	  28.8 &-11.2\\
\rule{0pt}{8pt} 0.5 &63.0	& 	28.4   &-10.5\\
\rule{0pt}{8pt} 0.7 &	60.7& 	 28.9  &-10.3\\
\rule{0pt}{8pt} 0.9 &	61.7& 	27.4   &-11.5\\

\bottomrule
\end{tabular}}
\caption{Performance comparisons of {\ouralg} (using T5-small as
the backbone) equipped with different $\gamma$ at task order 1.}
\label{tbl:hp_gamma}
\end{table}

About the selection of threshold for important skill units, the Table \ref{tbl:hp_delta} below shows the model's performance with varying thresholds $\delta$ on T5-small.

\begin{table}[t]
\centering
\scalebox{0.75}{
\begin{tabular}{lrrr}
\toprule
Importance Thresholds $\delta$  &Avg. JGA & FWT & BWT\\
\midrule
\rule{0pt}{4pt}1\% & 62.0 &26.3 &-9.4 \\
\rule{0pt}{8pt}5\%  &63.4 & 25.8& -10.1\\
\rule{0pt}{8pt}10\%  &62.2  &26.2&-9.5\\
\rule{0pt}{8pt}20\%  &62.1 &26.6&-9.1\\
\rule{0pt}{8pt}30\%  &62.7 &26.5&-10.0\\
\rule{0pt}{8pt}40\%  &60.9 &24.6&-10.2\\
\rule{0pt}{8pt}50\%  &54.8 &23.4&-10.3\\
\bottomrule
\end{tabular}}
\caption{Performance comparisons of {\ouralg} (using T5-small as
the backbone) equipped with different $\delta$.}
\label{tbl:hp_delta}
\end{table}

It can be seen that setting a high threshold (50\%) reduces model effectiveness by categorizing less significant skill units as important, which can contaminate historical knowledge and lead to forgetting. Conversely, a 1\% threshold still maintains strong performance owing to our effective skill consolidation approach, which effectively preserves task-specific knowledge and prevents forgetting. Considering that the heatmap in Figure \ref{fig:heatmap} displays approximately 20\% of the area in darker shades, signifying greater importance, we opted for a 20\% threshold to differentiate between important and unimportant skill units.

\section{Detailed Algorithm}
\label{sec:alg}
In this section, we provide the detailed implementation of {\ouralg} algorithm (see Algorithm~\ref{alg:my_algorithm}).

\begin{algorithm}[t]
\caption{{\ouralg}}\label{alg:my_algorithm}
\begin{algorithmic}[1]
\renewcommand{\algorithmicrequire}{\textbf{Input:}}
\renewcommand{\algorithmicensure}{\textbf{Output:}}
\REQUIRE Dataset $\mathcal{D}_k$ for task $k$ = $1,\ldots,K$; initial pre-trained model $f_0$; hyperparameters $\beta, \gamma$.

\STATE \textit{\# sequential tasks.}
\FOR {task $k$ = $1,\ldots,K$}  
    \STATE Get $f_k$ and calculate $\mathcal{U}_k$ by Algorithm (\ref{alg:ipt});
    \STATE Calculate $\delta_k$ based on $\mathcal{I}(\mathcal{U}_k)$;
    \IF{$k = 1$}
        \STATE \textit{\# initialization at beginning task.}
        \STATE $\hat{f}_1 \leftarrow f_1$, $\hat{\mathcal{U}}_1 \leftarrow \mathcal{U}_1$, $\hat{\delta}_1 \leftarrow \delta_1$;
    \ELSE
        \STATE \textit{\# fine-grained model averaging.} 
        \FOR {skill unit $i$ = $1,\ldots,n$}  
        \STATE Calculate $\hat{u}_i^k$ by (\ref{eq:5});
        \ENDFOR
        \STATE Get the averaged model $\hat{f}_k$ based on $\hat{\mathcal{U}}_k$;
        \STATE Calculate accumulated importance score $\mathcal{I}(\hat{\mathcal{U}}_k)$ according to (\ref{eq:norm});
        \STATE Calculate $\hat{\delta}_k$ based on $\mathcal{I}(\hat{\mathcal{U}}_k)$.
    \ENDIF   
\ENDFOR
\end{algorithmic} 
\end{algorithm}

\end{document}